\documentclass{article}

\usepackage[nonatbib,preprint]{nips_2018}

\usepackage[utf8]{inputenc} % allow utf-8 input
\usepackage[T1]{fontenc}    % use 8-bit T1 fonts
\usepackage{hyperref}       % hyperlinks
\usepackage{url}            % simple URL typesetting
\usepackage{booktabs}       % professional-quality tables
\usepackage{amsfonts}       % blackboard math symbols
\usepackage{nicefrac}       % compact symbols for 1/2, etc.
\usepackage{microtype}      % microtypography
\usepackage{graphicx}
\usepackage{tikz}
\usepackage{pgfplots}
\usepackage{array}
\title{Abstractive Text Classification Using Sequence-to-convolution Neural Networks}

\author{
  Taehoon~Kim~~~~~~Jihoon~Yang\\     
  Data Mining Research Laboratory \\
  Department of Computer Science and Engineering\\
  Sogang University\\
  \texttt{\{taehoonkim, yangjh\}@sogang.ac.kr} \\
}

\begin{document}

\maketitle

\begin{abstract}
We propose a new deep neural network model and its training scheme for text classification. Our model Sequence-to-convolution Neural Networks(Seq2CNN) consists of two blocks: Sequential Block that summarizes input texts and Convolution Block that receives summary of input and classifies it to a label. Seq2CNN is trained end-to-end to classify various-length texts without preprocessing inputs into fixed length.  We also present Gradual Weight Shift(GWS) method that stabilize training. GWS is applied to our model's loss function.  We compared our model with word-based TextCNN trained with different data preprocessing methods. We obtained significant improvement in classification accuracy over word-based TextCNN without any ensemble or data augmentation. Code is available at \href{https://github.com/tgisaturday/Seq2CNN}{https://github.com/tgisaturday/Seq2CNN}.  
\end{abstract}

\section{Introduction}

Ever since humans began to record information in the form of text, it was necessary to classify and manage information in a certain category to store and retrieve information efficiently. This need encouraged many researchers to develop a good text classification technique that can assign predefined categories to various kinds of text document such as emails, news articles, reviews, or patents. 

In commercial world, text classification techniques such as Naïve Bayes classifier\cite{CHEN20095432}, TFIDF\cite{Yun-tao2005}, Support Vector Machines(SVM)\cite{10.1007/BFb0026683} are already used in various fields including spam filtering, news categorization, and sentiment analysis. Recent development in deep neural networks\cite{DBLP:journals/corr/Kim14f,DBLP:journals/corr/ZhouSLL15b,AAAI159745,DBLP:journals/corr/ZhangZL15,DBLP:journals/corr/ConneauSBL16} are also achieving excellent results in extracting information from a text and classifying it into certain classes. 

As Convolutional Neural Networks(CNNs) achieved remarkable results in computer vision\cite{DBLP:journals/corr/SimonyanZ14a,DBLP:journals/corr/SzegedyIV16,DBLP:journals/corr/HeZRS15,DBLP:journals/corr/RenHG015}, researchers also applied CNNs to text classification\cite{DBLP:journals/corr/Kim14f,AAAI159745,DBLP:journals/corr/ZhangZL15,DBLP:journals/corr/ConneauSBL16} and showed excellent results. Training CNNs on top of pretrained word vectors\cite{DBLP:journals/corr/MikolovSCCD13,DBLP:journals/corr/JoulinGBM16,pennington2014glove} or character-level features\cite{DBLP:journals/corr/ZhangZL15,DBLP:journals/corr/ConneauSBL16} with hyperparameter tuning, they could get similar or outperforming results compared to other text classification models. 

Although TextCNNs’ performance in text classification is remarkable, they can only be applied to data whose input has fixed size. Since the number of parameters in TextCNN is determined by the length of input text, researchers had to crop or pad input texts into a certain length to train their TextCNN. This can result information loss when classifying longer texts and cause performance degradation.

In section 4.2, we show that performance of TextCNNs can be improved by training the model with summaries of input text. There are two ways to generate the summary of a text. One is extractive summarization, mere selection of a few existing sentences extracted from the source. The other is abstractive summarization, compressed paraphrasing of main contents of source, potentially using vocabulary unseen in the source. Both methods can change texts of various lengths into texts of fixed length still maintaining important features of source texts. 

TextRank\cite{mihalcea-tarau:2004:EMNLP} is a graph-based ranking model for extractive text summarization. TextRank gives a ranking over all sentences in a text allowing it to extract very short summaries without any training corpora. TextRank is widely used in summarizing structured text like news articles. 

Many researchers worked with Sequence-to-sequence Recurrent Neural Networks (Seq2seq RNNs)\cite{NIPS2014_5346,DBLP:journals/corr/NallapatiXZ16} to model abstractive text summarization. Using attention mechanism\cite{NIPS2015_5847} that allows neural networks to focus on different parts of their input, Seq2seq RNNs have been showing significant results in the task of abstractive summarization\cite{DBLP:journals/corr/RushCW15,DBLP:journals/corr/NallapatiXZ16}.

In this paper, we introduce Sequence-to-convolution Neural Networks(Seq2CNN) model that consists of two blocks: Sequence Block and Convolution Block. Sequence Block based on Attentional Encoder-Decoder Recurrent Neural Networks\cite{DBLP:journals/corr/NallapatiXZ16} summarizes input texts and feeds them into Convolution Block. Convolution Block based on TextCNN\cite{DBLP:journals/corr/Kim14f} classifies input texts into certain classes using the summaries provided by Sequential Block. Both blocks share non-static word embedding layer, encouraging them to collaborate for performance improvement. 

Simply connecting two blocks and train them with single end-to-end procedure cannot guarantee optimal results because Sequential Block doesn’t generate proper summaries in early stages of training. To solve this problem, we also propose a new training scheme that gradually shifts from fine-tuning for summarization task to fine-tuning for classification task as training progresses. Our model is implemented with Tensorflow\cite{tensorflow2015-whitepaper}. Code is available at \href{https://github.com/tgisaturday/Seq2CNN}{https://github.com/tgisaturday/Seq2CNN}. 

\section{Related Work}

There was similar approach of text classification with summaries using Latent Semantic Analysis(LSA) as extractive summarization method\cite{articleLSA}. They proposed a hybrid model for unlabeled document classification using SVM classifier with classification rules are generated using summaries of the training documents. Although we cannot directly compare the performance because of the domain difference in training data, we discuss performance of TextCNN trained with extractive summaries generated with TextRank\cite{mihalcea-tarau:2004:EMNLP}.  

Mnih et al.\cite{DBLP:journals/corr/MnihHGK14} came up with novel RNN models with visual attention that is capable of extracting information from an image or video by adaptively selecting a sequence of regions and this idea of using attention mechanism was successfully applied to machine translation by Bahdanau et al.\cite{DBLP:journals/corr/BahdanauCB14}. We used Bahdanau Attention\cite{DBLP:journals/corr/BahdanauCB14} to improve the performance of our Sequential Block. 

Our approach to use Attentional Encoder-Decoder RNNs for abstractive summarization is closely related to Nallapati et al.\cite{DBLP:journals/corr/NallapatiXZ16} who were the first to use Seq2seq RNNs and Attention model for abstractive text summarization. Our Tensorflow\cite{tensorflow2015-whitepaper} implementation of Sequential Block is very similar to Pan and Liu\cite{tensorflow-seq2seq}. They used Annotated English Gigaword\cite{Napoles:2012:AG:2391200.2391218} dataset to train their model and achieved state-of-the-art results (as of June 2016).

Seq2seq models\cite{NIPS2014_5346,DBLP:journals/corr/NallapatiXZ16} require training set of input text and corresponding example summary. None of datasets that we used for evaluation has any example summary. We used TextRank\cite{mihalcea-tarau:2004:EMNLP} to generate extractive summaries out of input texts and feed them to Sequential Block as example summary. We chose TextRank\cite{mihalcea-tarau:2004:EMNLP} because it can generate practical summaries even with short texts and doesn’t require any training corpora.

The architecture of our Convolution Block is based on TextCNN model proposed by Kim\cite{DBLP:journals/corr/Kim14f}. Kim\cite{DBLP:journals/corr/Kim14f} was the first to use CNN in text classification. Using this model as a baseline, we applied Batch Normalization after each convolution layer and changed hyperparameters for optimization.  

Our training scheme is closely related to the one proposed in Faster R-CNN paper\cite{DBLP:journals/corr/RenHG015}. Their scheme alternates between fine-tuning two different tasks. We also tried to train our model with training scheme that alternates between fine-tuning summarization task and classification task.However, this was not effective because summaries generated in early stages of training were filled with series of UNK (unknown word) tokens. Instead, our training scheme gradually changes focus of training from summary generation to text classification so that summaries generated in early stages of training don’t lower the classification accuracy of Convolution Block.

\section{Seq2CNN Model}
Figure 1 depicts the overall structure of Seq2CNN model.
\begin{figure}[!ht]
  \centering
  \includegraphics[width=\textwidth]{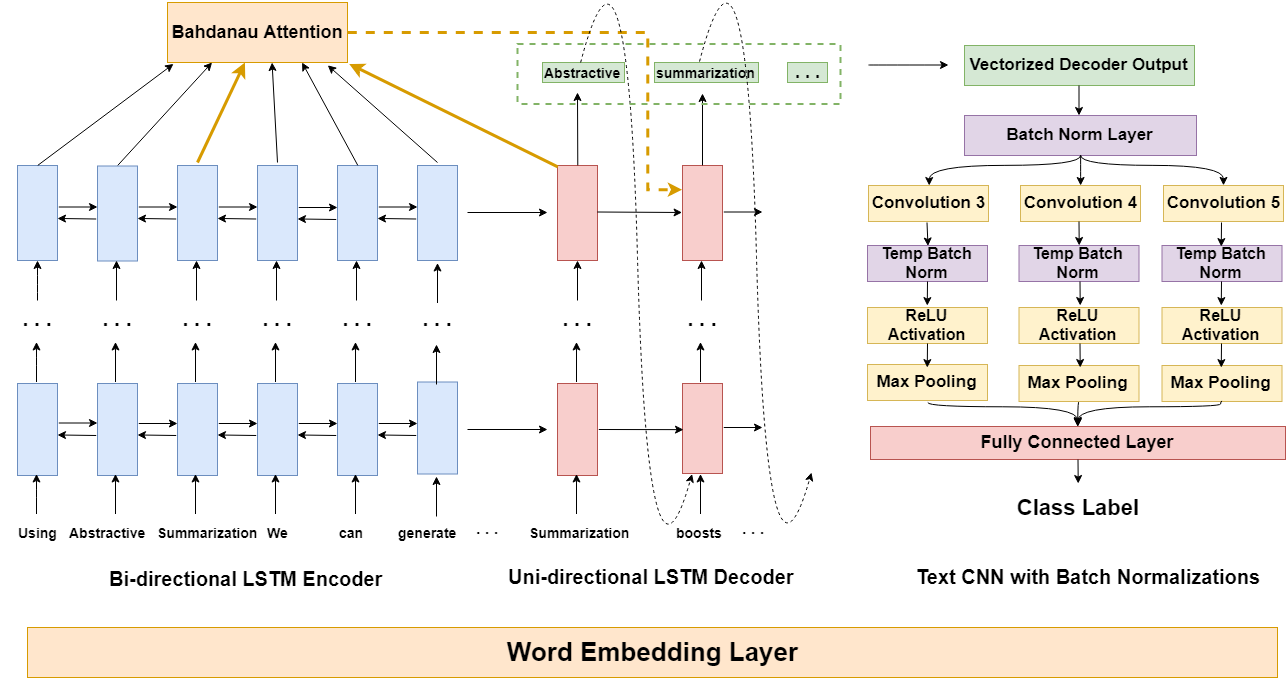}
  \caption{Overview of our Seq2CNN model that consists of Sequential Block(left) and Convolution Block(right). Both blocks interact with Word Embedding Layer to get vectorized representation of words used in summarization and classification tasks.}
\end{figure}

\subsection{Sequential Block}
Our baseline model of Sequential Block corresponds to the Attentional Encoder-Decoder RNN model used in Nallapati et al.\cite{DBLP:journals/corr/NallapatiXZ16} which encodes a source sentence into a fixed-length vector from which decoder generates abstractive summaries. The encoder consists of Bidirectional RNN\cite{650093} and decoder consists of Uni-directional RNN\cite{6638947}. We used Long Short-Term Memory RNN\cite{Hochreiter:1997:LSM:265493.264179} with 128 hidden units for encoder and decoder and Bahdanau Attention\cite{DBLP:journals/corr/BahdanauCB14} for attention mechanism. We also inserted Dropout\cite{JMLR:v15:srivastava14a} modules between LSTM layers to regularize.

The forward LSTM of encoder reads the input sequence as it is ordered, and the backward LSTM reads the sequence in the reverse order. In this way, fixed-length vector from encoder contains the summaries of both preceding words and the following words. With the help of attention mechanism, decoder decides parts of the source sentence to pay attention to and focus only on the vectors that are essential for summarization. 

\subsection{Convolution Block}
The structure of our Convolution Block is based on TextCNN model proposed by Kim\cite{DBLP:journals/corr/Kim14f} which gets $n \times k$ vectorized representation of text where $n$ is the number of words inside the text and $k$ is the dimension size of word embedding. Each filter windows with varying sizes($h$) extracts one feature by performing $h \times k$ convolution operation over input and apply max-over-time pooling operation. The model uses multiple filters to multiple features. These features are passed to a fully-connected softmax layer whose output is the probability distribution over labels.

Convolution Block gets vectorized representation of summaries generated by Sequential Block. We used rectified linear units(ReLU)\cite{DBLP:journals/corr/XuWCL15} for non-linear activation function and filter windows of 3, 4, 5 with 32 filters each. We applied Batch Normalization\cite{Ioffe:2015:BNA:3045118.3045167} after each convolution layer. Batch Normalization\cite{Ioffe:2015:BNA:3045118.3045167} accelerates training by reducing internal covariate shift\cite{Ioffe:2015:BNA:3045118.3045167}, the change in the distribution of network activations due to the change in network parameters during the training. We also applied batch normalization on vectorized representation of summaries to stabilize entire training procedure by reducing internal covariate shift between Sequential block and Convolution block. For regularization, we inserted Dropout\cite{JMLR:v15:srivastava14a} module between max-pooling layer and fully-connected layer.

\subsection{Word Embedding Block}
Word Embedding Block consists of word embedding layer which stores vectorized representations of each word and vocabulary lookup table that maps each word with corresponding vector representation. To make vocabulary dictionary, we extracted 20,000 to 30,000 words from training data with minimum word frequency($f$), excluding words that have appeared less than $f$ times. Our word embedding layer is non-static and fine-tuned via back-propagation. In our implementation, we set word embedding dimension to 100.

\subsection{Loss Function}
The main objective of Seq2CNN is to classify texts of various lengths without losing important features of original context. For feature extraction, we use abstractive summarization method using Sequential Block. Although our model is mainly focused on classification, quality of summary must be guaranteed for Convolution Block to successfully perform classification task. 

Taking everything into consideration, we trained our model to minimize an objective function which is weighted sum of losses in classification and summarization. Our loss function for a text is defined as: 
\begin{equation}
L(\{p_{y_i},y_i^{*}\},\{t_i,t_i^*,l_i\})\; = \; \frac{1}{N}\left [ L_{cls}(p_{y_i},y_i^{*})\; +\, \lambda L_{sum}(t_i,t_i^*,l_i) \right ]
\end{equation}

\begin{equation}
L_{sum}(t_i,t_i^*,l_i)\; = \; \frac{1}{l_{i}}\sum_{j}^{l_{i}}L_{vocab_j}(p_{w_j},w_j^*)
\end{equation}
In (1), $i$ is the index of a text in a mini-batch of size $N$ and $p_{y_i}$ is the predicted probability of text $i$. classified as ground truth label $y_i^*$. The classification loss $L_{cls}$ is cross-entropy loss over classification classes. The summarization loss $L_{sum}$ is sequence loss between the summary output of Sequential Block $t_i$ with length $l_i$ and summary example $t_i^*$. 

In (2), $p_{w_j}$ is the predicted probability of the $j$th word $w_j$ in generated summaryto match with the $j$th word of summary example $w_{j^*}$. We defined sequence loss as the average of the vocabulary losses $L_{vocab}$s in $t_i$. The vocabulary loss of $j$th word $L_{vocab_j}$ is cross-entropy loss over vocabulary in dictionary stored in Word Embedding Block. 

Total loss is weighted sum of $L_{cls}$ and $L_{sum}$ with a balancing weight $\lambda$, normalized with $N$. In our implementation, we used multi-class softmax layer to get $p_{y_i}$ and $p_{w_j}$. We normalized our loss with the mini-batch size. We applied Gradual Weight Shift to $\lambda$. We explained more about Gradual Weight Shift in section 3.5. 

\subsection{Gradual Weight Shift}
In our implementation of Seq2CNN, we train Sequential Block and Convolution Block end-to-end by back-propagation and gradient optimizer using loss function defined in 3.4. This training scheme fine-tunes the model and reduces training time compared to training each model independently. However, using a constant value (\textgreater1.0) for balancing weight $\lambda$ caused sudden drops of validation accuracy in later stages of training (\textasciitilde20 epochs). We found the main cause of this phenomenon in Sequential Block.

In earlier stages of training (\textasciitilde10 epochs), Sequential Block does not generate practical summaries and omits UNK tokens instead. Huge difference in quality of summary throughout the training hinders the optimization of Convolution Block. Using larger $\lambda$ (\textgreater1.5) solves this problem a little bit by giving more weight to $L_{sum}$, making Sequential Block to converge faster.

Using larger $\lambda$ gives more weight to $L_{sum}$ than $L_{cls}$, leading the model to focus on optimization of Sequential Block until the end. Since our model is designed for classification, we gradually shifted weight from $L_{sum}$ to $L_{cls}$ by exponentially decaying $\lambda$ throughout time. 
Our exponential decay function for $\lambda$ is defined as:
\begin{equation}
\lambda_{t}\; =\; \lambda_{0}\delta ^{t}
\end{equation}
where $\lambda_0$ is initial value of lambda, $\delta$ is decay rate, and $t$ is current time step. We could stabilize the training of our model and achieve higher test accuracy by applying Gradual Weight Shift to our loss function defined in section 3.4. We explain more about the result in section 4.4. 

\subsection{Sharing Word Embedding for Summarization and Classification}
We designed our model to share word embedding for summarization and classification. When back-propagation happens, Sequential Block tries to update word embedding layer in direction of minimizing sequence loss. On the other hand, Convolution Block tries to update word embedding layer in direction of minimizing classification loss. This fine-tunes word embedding layer. 

\subsection{Optimization}
In our implementation, we trained our model end-to-end by back-propagation and stochastic gradient descent(SGD)\cite{SGD} using Adam optimizer\cite{DBLP:journals/corr/KingmaB14} with $\beta_{1}=0.9$, $\beta_{2}=0.999$, and $\epsilon = 0.1$. We used learning rate of 0.001, decayed every epoch using an exponential rate of 0.95. Dropout rate for Dropout\cite{JMLR:v15:srivastava14a} modules is 0.7. We also used gradient clipping\cite{DBLP:journals/corr/abs-1211-5063} with gradient norm limited to 5. We set loss balancing weight $\lambda$ to 1.0 when training AG's News and 2.0 for other datasets.

We initialized convolution layers using He Normal\cite{DBLP:journals/corr/HeZR015} initializer and fully-connected layer using Xavier\cite{pmlr-v9-glorot10a} initializer. All other weights were initialized with random values from a uniform distribution with minimum of -1 and maximum of 1. The implementation is done using Tensorflow\cite{tensorflow2015-whitepaper}. We trained our model using single NVIDIA Titan V GPU with mini-batch size 256. It took 3 hours(AG’s News) to 1 day(Yahoo Answers) for training. We didn't used any ensemble or data augmentation techniques Detailed information about datasets are given in section 4.1.

\section{Experiments}
We evaluated the performance of our model by comparing with basic TextCNN\cite{DBLP:journals/corr/Kim14f}. We defined basic TextCNN used in experiments as Vanilla CNN. We used the same hyperparameters(filter windows of 3, 4, 5 with 32 filters each) for Vanilla CNN and Convolution Block in Seq2CNN. We trained Vanilla CNN with three different data preprocessing methods: \textbf{full-text, crop\&pad, summarize}. We defined the length of text as the number of words in each text. 
\paragraph{full-text}
This is default data preprocessing method. We just removed unnecessary characters and stopwords from the input texts and padded them into fixed length with PAD token. Here, fixed length is the maximum length of text in each dataset. Same preprocessing method is also applied before \textbf{crop\&pad} and \textbf{summarize}.
\paragraph{crop\&pad}
Using fixed length as hyperparameter(numbers inside bracket in Table 2), we cropped each text into fixed length. For example, if a text is longer than 20 words, we only used 20 words starting from the front. Texts shorter than fixed length is padded with PAD token.
\paragraph{summarize}
Instead of cropping each sentence, we generated extractive summary of input text using TextRank\cite{mihalcea-tarau:2004:EMNLP} and processed the summary with \textbf{crop\&pad}. Same method was used to generate summary example used in training Sequential Block.  
\subsection{Datasets}
\begin{table}[!ht]
  \caption{Statisics of datasets. Vocabulary size is number of words used to train the model. Min Freq is minimum word frequency $f$ used to decide vocabulary size of each dataset.}
  \label{table1}
  \centering
  \begin{tabular}{lccccc}
    \toprule
    Datasets & Classes& Training Set&Test Set & Vocabulary Size& Min Freq\\
    \midrule
   AG's News &4  &120,000 &7,600  &26,543 &3  \\
    DBPedia  &14 &140,000 &70,000 &23,371 &15 \\
    Yahoo Answers &10 &150,000 & 60,000  &28,451 &15  \\
    \bottomrule
  \end{tabular}
\end{table}
We evaluated our model on three different datasets: AG's News, DBPedia, and Yahoo Answers\cite{DBLP:journals/corr/ZhangZL15}. To offer fair evaluation on the performance of Sequential Block in Seq2CNN, we removed input texts shorter than 50 words(maximum fixed length value in Table 2), changing number of training samples in DBPedia and Yahoo Answers. The number of training samples in AG' News is the same as the original. We didn't apply any changes to test samples for fair comparison with best published results. We also limited the size of vocabulary into 20,000 to 30,000 with minimum word frequency($f$) in Table 1. Words in test samples are not included in vocabulary dictionary of Word Embedding Block. None of the datasets contains summary examples to train Seq2CNN model so we generated summary examples using TextRank\cite{mihalcea-tarau:2004:EMNLP}. We didn't feed any summary samples while evaluating Seq2CNN with test samples.\footnote{In our implementation with Tensorflow, we used greedy embedding helper instead of training helper for inference layer.}  Detailed statistics of each dataset is given in Table 1. 

\subsection{Text Classification}
\begin{table}[!ht]
  \caption{Classification results of all models. Numbers are test accuracy in percentage. "Vanilla CNN" is basic TextCNN\cite{DBLP:journals/corr/Kim14f} model and "Seq2CNN" is our model. "Full" stands for "full-text", "Crop" stands for "crop\&pad", and "Sum" stands for "summarize". We labeled the best result of Vanilla CNN in blue and worst result in red. Best result of Seq2CNN is labeled in green.}
  \label{table3}
  \centering
  \begin{tabular}{lccc}
    \toprule
    Model & AG's News & DBPedia &Yahoo Answers\\
    \midrule
    Vanilla CNN Full &\textcolor{red}{89.28}  &96.39 &54.37  \\
    Vanilla CNN Crop(20)  &89.42	&96.69	&54.28  \\
    Vanilla CNN Crop(50)  &89.48	&\textcolor{red}{96.38}	&\textcolor{blue}{54.83}  \\
    Vanilla CNN Sum(20)  &89.59	&96.84	&\textcolor{red}{53.98}  \\
    Vanilla CNN Sum(50)  &\textcolor{blue}{89.67}	&\textcolor{blue}{96.89}	&54.69 \\
    Seq2CNN(20) &90.18	&97.07	&55.06 \\
    Seq2CNN(50) &\textcolor{green!55!blue}{90.36}	&\textcolor{green!55!blue}{97.23}	&\textcolor{green!55!blue}{55.39} \\
    \bottomrule
  \end{tabular}
\end{table}

\paragraph{Extractive Text Classification}
Table 2 shows classification results of Seq2CNN and Vanilia CNN models. We first evaluated data preprocessing methods for Vanilla CNN with different fixed length sizes. We labeled the best result in blue and worst result in red. There was not a single data preprocessing method that derived best performance for all datasets. Summarization with TextRank\cite{mihalcea-tarau:2004:EMNLP} tends to work well in most of the cases.     
\paragraph{Abstractive Text Classification}
Our Seq2CNN model outperformed other models in all cases bringing average 1\% growth compared to Vanilla CNN trained without any data preprocessing(Vanilla CNN Full). Best published result for AG's News using TextCNN\cite{DBLP:journals/corr/ZhangZL15} is 90.09\% with pretrained word2vec\cite{DBLP:journals/corr/MikolovSCCD13} embedding and data augmentation technique\cite{DBLP:journals/corr/ZhangZL15} using thesaurus. Our model achieved competitive result on AG's News dataset without any pretrained word embedding or data augmentation technique. We cannot directly compare other results due to the changes that we explained in section 4.1. 
\subsection{Text Summarization}
\begin{table}
  \caption{Examples of output produced by Sequential Block with short-length texts. TextRank failed to generate any summary, returning the first sentence of input instead.}
  \label{table2}
  \centering
  \setlength{\extrarowheight}{2pt}
  \begin{tabular}{l|l|p{8cm}}
    \toprule
    Type & Label & Sentence \\
    \midrule
    Original & Sports & \parbox{8cm}{Great Britain's Amir Khan, who looked so impressive in winning the 132-pound championship at the Junior International Invitational Boxing Championships here last summer, has a chance for an Olympic gold medal in the lightweight division today.} \\
    \hline
    TextRank  & Sports &\parbox{8cm}{Great Britain's Amir Khan, who looked so impressive in winning the 132-pound championship at the Junior International Invitational Boxing Championships here last summer,}\\
    \hline
    Sequential Block  &Sports & \parbox{8cm}{great britain amir khan looked impressive winning pound championship junior international invitational boxing championships last summer chance olympic gold medal lightweight division today} \\
    \hline
    \hline
    Original & Sci/Tech & \parbox{8cm}{A ROBOT that will generate its own power by eating flies is being developed by British scientists. The idea is to produce electricity by catching flies and digesting them in special fuel cells that will break} \\
    \hline
    TextRank  & Sci/Tech &\parbox{8cm}{A ROBOT that will generate its own power by eating flies is being developed by British scientists.}\\
    \hline
    Sequential Block  &Sci/Tech & \parbox{8cm}{robot generate power eating flies developed british scientists idea produce electricity catching flies digesting special fuel cells break} \\
    \bottomrule
  \end{tabular}
\end{table}
TextRank\cite{mihalcea-tarau:2004:EMNLP} algorithm cannot generate proper summary if the original text is too short. As a result, classification with Vanilla CNN and TextRank(Vanilla CNN Sum(20)) performed worst with Yahoo Answers dataset.\footnote{We designed our implementation of TextRank algorithm to return the first sentence if the original text was too short.} Seq2CNN is robust to short-length texts as it's shown in Table 3. Even with short-length texts, Sequential Block successfully generate summaries by removing unimportant words from the original. TextRank\cite{mihalcea-tarau:2004:EMNLP} algorithm failed to generate any summary for both examples. 

\subsection{Gradual Weight Shift}
\begin{figure}
  \centering
\begin{tikzpicture}
\begin{axis}[
    axis lines = left,
    xlabel = $Steps$,
    ylabel = {$Loss$},
	title={Total Loss},
	width=0.5\linewidth,
	no marks]
\addplot[line width=1pt,smooth,color=red] %
	table[x=Step,y=Value,col sep=comma]{wo_GWS_loss.csv};
\addlegendentry{without GWS};
\addplot[line width=1pt,smooth,color=blue] %
	table[x=Step,y=Value,col sep=comma]{w_GWS_loss.csv};
\addlegendentry{with GWS};
\end{axis}
\end{tikzpicture}
\begin{tikzpicture} 
\begin{axis}[
    axis lines = left,
    xlabel = $Steps$,
    ylabel = {$Loss$},
	title={Sequence Loss},
	width=0.5\linewidth,
	no marks]
\addplot[line width=1pt,smooth, color=red] %
	table[x=Step,y=Value,col sep=comma]{w_GWS_seq_loss.csv};
\addlegendentry{without GWS};
\addplot[line width=1pt,smooth,color=blue] %
	table[x=Step,y=Value,col sep=comma]{wo_GWS_seq_loss.csv};
\addlegendentry{with GWS};
\end{axis}
\end{tikzpicture}
  \caption{Total loss curve(left) and sequence loss curve(right) on the \textbf{AG's News Dataset} with and without Gradual Weight Shift(GWS). }
\end{figure}
\begin{table}[ht]
  \caption{Classification results on the \textbf{AG's News Dataset} with and without Gradual Weight Shift(GWS). Best result using Vanilla CNN is also included for comparison.}
  \label{table3}
  \centering
  \begin{tabular}{lc}
    \toprule
    Model & Accuracy\\
    \midrule
    Vanilla CNN  &89.67 \\
    Seq2CNN without GWS(20)  &89.75  \\
    Seq2CNN without GWS(50)  &89.87	  \\
    Seq2CNN with GWS(20)  &90.18 \\
    Seq2CNN with GWS(50)  &90.36  \\
    \bottomrule
  \end{tabular}
\end{table}

\paragraph{Optimization}In sequence loss curve of Figure 2, the sequence loss of the model trained without GWS converges smoothly in the early stage of training, but starts to fluctuate after 2,000 steps. This also effects the total loss curve, unstabilizing the training of the model. In contrast, loss curves of the model with GWS converges smoothly until the end. 

\paragraph{Classification Performance}We evaluated the performance of Gradual Weight Shift with AG's News Dataset. In Table 4, Seq2CNN model trained with GWS achieved better results compared to the same model trained without GWS. Although Seq2CNN model without GWS performed better than any other Vanilla CNN models, it could not outperform previous best published result of Zhang et al.\cite{DBLP:journals/corr/ZhangZL15}.   

\section{Conclusion}
We have proposed Sequence-to-Convolution Neural Networks (Seq2CNN) for efficient and accurate text classification. Seq2CNN can be trained with texts of various lengths without any text preprocessing method such as cropping or summarizing. We also presented a new training theme for our model using Gradual Weight Shift(GWS), which can be applied to other models with multi-task loss function by changing number of balancing weights.

The true strength of Seq2CNN comes from its flexibility. Each blocks can be replaced with other models designed for the same tasks. For example, Sequential Block can be replaced with multi-layer Seq2Seq model\cite{tensorflow2015-whitepaper} or Text Variational Autoencoder\cite{DBLP:journals/corr/SemeniutaSB17}. Convolution Block can be replaced with other text classification models such as C-LSTM\cite{DBLP:journals/corr/ZhouSLL15b}, Recurrent-CNN\cite{AAAI159745}, Char-CNN\cite{DBLP:journals/corr/ZhangZL15}, or VDCNN\cite{DBLP:journals/corr/ConneauSBL16}.

We adopted general sequence loss function which uses predicted probability of $j$th word $w_j$ in generated summary matches $j$th word of summary example $w_{j^*}$ to calculate vocabulary losses. However, we think the sequence loss cannot be evaluated accurately by comparing only the words in the same position. Bahdanau et al.\cite{DBLP:journals/corr/BahdanauSBKCCB15} suggested specialized surrogate losses for Encoder-Decoder models often used for sequence prediction tasks and brought significant performance improvements.

We also didn't use any pretrained word embeddings to initialize the Word Embedding Block. Previous results on word-based TextCNN \cite{DBLP:journals/corr/Kim14f,DBLP:journals/corr/ZhangZL15,AAAI159745} suggests that initializing embedding layers with pretrained word vectors such as word2vec\cite{DBLP:journals/corr/MikolovSCCD13}, FastText\cite{DBLP:journals/corr/JoulinGBM16}, or GloVe\cite{pennington2014glove} helps improves performances of models.

In the future, we are planning to improve Seq2CNN by reassembling our model with every possible methods.

%\bibliography{mybib}{}
%\small
%\bibliographystyle{plain}

\end{document}